\documentclass{article}
\usepackage{amssymb}
\usepackage{booktabs}
\usepackage{multirow} 
\usepackage{tabularx}
\usepackage[table]{xcolor}
\usepackage{bbm}
\usepackage{graphicx}
\usepackage{amsmath}
\usepackage{cuted}
\usepackage{capt-of}
\usepackage{float}
\usepackage{pgfplots}
\usepackage{soul}
\usepackage{pgfplotstable}
\pgfplotsset{compat=newest}

\newcommand{\acronym}{\texttt{NeME}}
\newcommand{\acronymnorm}{NeME}
\usetikzlibrary{intersections}
\usepgfplotslibrary{fillbetween}

\usepackage[preprint]{corl_2025} 

\title{Learning to Evaluate Autonomous Behaviour in Human-Robot Interaction}

\author{
  Matteo Tiezzi, Tommaso Apicella, Carlos Cardenas-Perez, Giovanni Fregonese, \\ \textbf{Stefano Dafarra, Pietro Morerio, Daniele Pucci, Alessio Del Bue}\\
 Istituto Italiano di Tecnologia (IIT), Genoa, Italy\\
  \texttt{firstname.secondname@iit.it} \\
}

\begin{document}
\setul{0.3ex}{0.6ex}

\makeatletter
\g@addto@macro\@maketitle{
  \begin{figure}[H]
  \setlength{\linewidth}{\textwidth}
  \setlength{\hsize}{\textwidth}
    \centering
  \begin{tikzpicture}

\begin{axis}[
    width=0.87\linewidth,
    height=0.4\linewidth,
    xlabel={Epoch},
    ylabel={Loss value},
     ymin=0, ymax=1.2,
    xmin=1, xmax=12,
    legend style={at={(0.32,1.)}, anchor=south, legend columns=2, font=\small, draw=none},
    ylabel shift = -5pt,
    grid=major,
    axis y line*=left,
    axis x line*=bottom,
    ytick={0, 0.2, 0.4, 0.6, 0.8, 1, 1.2},
    yticklabels={0, 0.2, 0.4, 0.6, 0.8, 1, 1.2},   
    grid style={line width=.1pt, draw=gray!30},
    label style={font=\small},
    tick label style={font=\small}
]


\addplot[
    color=gray!80,
    solid,
    line width=1.5pt,
    mark=triangle*,
    mark size=1.5pt,
] coordinates {
(1,0.877) (2,0.936) (3,0.927) (4,0.961) (5,0.96) (6,0.953) (7,0.966) (8,0.984) (9,0.992) (10,0.97) (11,0.985) (12,1.04)
};

\addlegendentry{Validation Loss}

\end{axis}

\begin{axis}[
    width=0.87\linewidth,
    height=0.4\linewidth,
    xlabel={Epoch},
    ylabel={DTW},
    ymin=2.0, ymax=2.8,
    xmin=1, xmax=12,
    axis y line*=right,
    axis x line=none,
    legend style={at={(0.7,1.0)}, anchor=south, legend columns=2, font=\small, draw=none},
    label style={font=\small},
    tick label style={font=\small}
]
\pgfplotsset{every outer y axis line/.style={xshift=1.3cm}, every tick/.style={xshift=1.3cm}, every y tick label/.style={xshift=1.3cm} }

\addplot[
    color=cyan!60,
    line width=1.5pt,
    mark=square*,
    mark size=1.5pt,
] coordinates {
(1,2.83) (2,2.59) (3,2.41) (4,2.33) (5,2.29) (6,2.27) (7,2.18) (8,2.18) (9,2.16) (10,2.13) (11,2.20) (12,2.10)
};
\addlegendentry{Dynamic Time Warping (DTW)}
\end{axis}

\begin{axis}[
    width=0.87\linewidth,
    height=0.4\linewidth,
    xlabel={Epoch},
    ylabel={SR / \textbf{\acronymnorm} (mF1)},
    ymin=20, ymax=100,
    xmin=1, xmax=12,
    axis y line*=right,
    axis x line*=none,
    ytick={20, 40, 60, 80, 100},
    yticklabels={20, 40, 60, 80, 100},  
    legend style={at={(0.5,1.1)}, 
    anchor=south, legend columns=2, font=\small, draw=none},
    label style={font=\small},
    tick label style={font=\small},
    ylabel style={yshift=7pt, xshift=0pt},
]


\addplot[
    color=blue!60,
    line width=1.5pt,
    mark=diamond*,
    mark size=1.5pt,
] coordinates {
(1,33.33) (2,55.83) (3,57.5) (4,66.67) (5,51.67) (6,75) (7,75) (8,70) (9,63.5) (10,57.5) (11,68.33) (12,66.6)
};
\addlegendentry{Average Success Rate (SR)}

\addplot[
    color=brown!60,
    dashed,
    line width=3pt,
    mark=o,
    mark size=2pt,
] coordinates {
(1,63.59) (2,68.51) (3,70.4) (4,68.57) (5,57.47) (6,68.06) (7,67.32) (8,71.25) (9,70.6) (10,58.25) (11,62.83) (12,66.6)
};
\addlegendentry{\textbf{\acronymnorm} (mF1)}

\addplot[name path=upper, draw=none] coordinates {
(1,63.50) (2,87.19) (3,80.36) (4,82.44) (5,71.95) (6,99.85) (7,91.93) (8,89.49) (9,82.48) (10,89.68) (11,90.68) (12,89.93)
};

\addplot[name path=lower, draw=none] coordinates {
(1,3.16) (2,24.47) (3,34.64) (4,50.90) (5,31.39) (6,50.15) (7,58.07) (8,50.51) (9,44.52) (10,25.32) (11,45.98) (12,43.27)
};

\addplot[blue!20, fill opacity=0.5] fill between[of=upper and lower];
\end{axis}

\end{tikzpicture}
  \vspace{-7pt}
  \caption{
  Model selection in Imitation Learning is challenging. The success rate is the most used metric, but requires time-consuming robot deployment for each epoch, is sensitive to variability (standard deviation among three subjects), and does not consider the robot movement. The validation loss  does not identify the best model, since it represents only a surrogate for the the task success rate. Dynamic Time Warping can be used as a measure for scoring the best model computing the alignment between the predicted and annotated joints positions over time, however similar trajectories in different space locations might have different DTW. In contrast, our method \acronym{} considers the movement of the joints, and enables efficient offline model selection and testing without robot-deployment closely aligning with the model chosen using success rate.
  }
\label{fig:main_figure}
\end{figure}
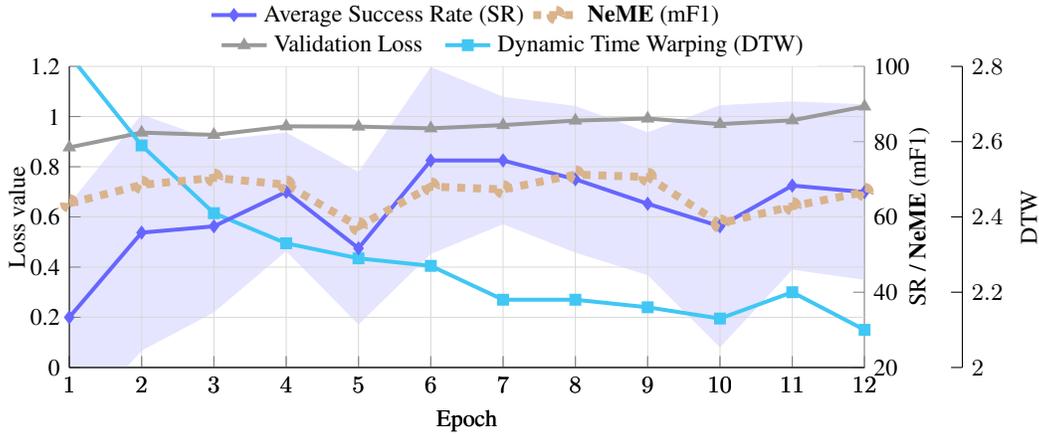
}
\makeatother
\maketitle


\begin{abstract}
Evaluating and comparing the performance of autonomous Humanoid Robots is challenging, as success rate metrics are difficult to reproduce and fail to capture the complexity of robot movement trajectories, critical in Human-Robot Interaction and Collaboration (HRIC). To address these challenges, we propose a general evaluation framework that measures the quality of Imitation Learning (IL) methods by focusing on trajectory performance. We devise the Neural Meta Evaluator (NeME), a deep learning model trained to classify actions from robot joint trajectories. NeME serves as a meta-evaluator to compare the performance of robot control policies, enabling policy evaluation without requiring human involvement in the loop. We validate our framework on ergoCub, a humanoid robot, using teleoperation data and comparing IL methods tailored to the available platform. The experimental results indicate that our method is more aligned with the success rate obtained on the robot than baselines, offering a reproducible, systematic, and insightful means for comparing the performance of multimodal imitation learning approaches in complex HRI tasks. 
\end{abstract}



\section{Introduction}
\label{sec:introduction}
Imitation Learning (IL)~\cite{hussein2017imitation} has become a cornerstone of modern robotics, enabling agents to acquire complex behaviours by replicating expert demonstrations. Among these approaches, behavioural cloning (BC)~\cite{ross2010efficient} emerged as one of the primary paradigms~\cite{fosterbehavior}: it frames the generation of autonomous behaviours as a supervised learning problem, where a policy is trained to directly map sensory inputs—including visual observations and proprioceptive signals—to desired joint configurations provided by human operators, typically via teleoperation interfaces~\cite{cardenas2024xbg, zhang2018deep,fu2024humanplus,rahmatizadeh2018vision}. BC-trained methods have been successfully deployed across a wide spectrum of robotic platforms, ranging from robotic arms to humanoid systems, supporting diverse tasks such as manipulation~\cite{mizuchi2023designing,abbo2025concerns,coronado2022evaluating,bensch2017interaction,kokotinis2023quantification} and Human-Robot Interaction and Collaboration (HRIC)~\cite{cardenas2024xbg}. 

Despite its widespread adoption, assessing and comparing the performance of different imitation learning methods, particularly in HRIC scenarios, remains challenging. Actions performed by humans when interacting with the robot are inherently variable, challenging the evaluation of several trained policies under exactly identical conditions.
Existing evaluation approaches have mainly relied on coarse success rate metrics, which require human involvement in the evaluation loop~\cite{cardenas2024xbg,zhang2018deep,fu2024humanplus, rahmatizadeh2018vision, wang2023mimicplay, mandlekar2020learning, zhao2023learning,fu2024mobile,brohan2022rt,brohan2023rt,team2025gemini, bjorck2025gr00t} or on the alignment between the robot's joints trajectory and the movements of the human demonstrator~\cite{taghavi2021online,dietzel2024assessing}. 
However, success rates only fail to capture the nuanced qualities of the whole robot motion itself, 
while strict trajectory alignment penalizes robot behaviours that are correct but differ from demonstrations, potentially hindering human safety and task effectiveness in HRIC, especially with humanoid robots.
%
Moreover, direct evaluation on robots is time-consuming and resource-intensive: each model must be deployed under identical hardware/software conditions, while human variability and environmental factors further complicate fair comparisons. Although simulations can help, the sim-to-real gap remains a significant obstacle \cite{afzal2020study, gu2024humanoid, 10610977}. These difficulties are worsened during model selection, where evaluating numerous hyperparameter settings would require repeated real-world deployments. Such a step is necessary, as recent works \cite{NEURIPS2020_51200d29,paine2020hyperparameter} highlight that validation loss, although commonly used for model selection, may represent a surrogate for the true objective of interest, i.e. the task
success rate. We show this phenomenon in Figure~\ref{fig:main_figure}, where discrepancies between the loss curves and the success rate emerge.

To address these challenges, we propose an evaluation framework that measures the performance of trained policies focusing on the characteristics of the movement trajectories. In doing so, we defer the evaluation of trained IL policies to a mechanism that is learned itself, aiming at the removal of subjective and tailored protocols from the evaluation pipeline, mainly due to the involvement of humans in the evaluation loop. 
In particular, we devise a \textit{Neural Meta Evaluator} -- \acronym{}, trained to classify the action performed by the robot, based on the sequence of joints positions, i.e. the joints trajectory. Once trained, \acronym{} is employed as a \textit{meta-evaluator}--that is, a model capable of objectively assessing the performance of different imitation learning policies by analysing the produced trajectories. 
In our experiments, we analyse the performance and characteristics of several sequence processing models following recent literature~\cite{tiezzi2024state}, including state-space models and the latest variants of recurrent architectures, investigating their ability to correctly identify actions from robot joints positions across varying context lengths. We validate \acronym{} on different variants of eXteroceptive Behaviour Generation model (XBG)~\cite{cardenas2024xbg}, an architecture trained to work in HRIC applications. 
We show that our method serves as a complementary tool to traditional success rate evaluation (Figure~\ref{fig:main_figure}), enabling model selection based on data not seen during training and without the need for deployment on the physical robot. We validate this approach on the ergoCub~\cite{ergocub2023} robotic platform, showing the solution's ability to streamline and speed up the model selection and evaluation process. Furthermore, \acronym{} can be used to perform model selection during training, overcoming limitations where validation loss may not provide a reliable indicator of policy quality~\cite{NEURIPS2020_51200d29,paine2020hyperparameter,JustinFu,mandlekarmatters}.



\section{Related Works}
\label{sec:related}

\textbf{Learning behaviours with IL.} Imitation Learning (IL) refers to the process of learning the generation of autonomous behaviours from expert demonstrations~\cite{hussein2017imitation}. A primary approach within IL is Behavioural Cloning (BC)~\cite{ross2010efficient,pomerleau1988alvinn}, where the robot control problem is formulated as supervised learning.
In BC, methods (or policies)--typically modelled as neural networks--are trained to predict robot joint positions directly from sensory observations e.g., visual inputs, proprioception, and task information. 
Training supervision is provided by ground truth joint trajectories collected via teleoperation~\cite{cardenas2024xbg,zhang2018deep,fu2024humanplus,rahmatizadeh2018vision, zhao2023learning,team2025gemini,bjorck2025gr00t}. 
Most BC-based methods assume that task information is either known a prior or explicitly provided as input to the model, alongside sensory observations 
~\cite{zhang2018deep,fu2024humanplus,wang2023mimicplay,mandlekar2020learning,zhao2023learning,fu2024mobile,brohan2022rt,brohan2023rt,team2025gemini,bjorck2025gr00t,kim2024openvla}. When task knowledge is provided a priori, a separate model is typically trained for each task~\cite{fu2024humanplus,zhao2023learning,fu2024mobile}. Conversely, Vision-Language-Action (VLA) models~\cite{brohan2022rt, brohan2023rt,team2025gemini,bjorck2025gr00t,kim2024openvla} accept a textual description of the task as additional input. For instance,  HumanPlus~\cite{fu2024humanplus} trains separate models on each different task, selectively masking a joints subset of joints depending on the task; methods like Gr00t~\cite{bjorck2025gr00t} and Gemini Robotics~\cite{team2025gemini} encode the task using text tokens,  combining them with the proprioceptive and visual features. 
These approaches, however, require a prior on the task to be performed and are not able to generate joints positions purely in response of the behaviour of a human in the field of view--a critical capability for HRIC applications. On the contrary, eXteroceptive Behaviour Generation (XBG)~\cite{cardenas2024xbg} directly maps human-robot interaction data to robot joint trajectories, enabling context-aware robot behaviours based on the human movements. 

\textbf{Evaluating imitation learning policies.} 
A common 
evaluation strategy for BC-trained models is the use of
success rate metrics~\cite{cardenas2024xbg,zhang2018deep,fu2024humanplus,rahmatizadeh2018vision, wang2023mimicplay, mandlekar2020learning, zhao2023learning, fu2024mobile,brohan2022rt, brohan2023rt,team2025gemini, bjorck2025gr00t}, measuring the percentage of completed tasks.  
However, success rate evaluation presents two main limitations.
%
First, it does not capture the \textit{quality} of the generated motion.
Most of the works training models with BC evaluate the performance of methods on robotic arms~\cite{zhang2018deep, rahmatizadeh2018vision, wang2023mimicplay, mandlekar2020learning, zhao2023learning, fu2024mobile,brohan2022rt, brohan2023rt} in tasks such as reaching, pushing, picking and placing an object, without involving interactions with a human and with relatively low-dimensional control. In contrast, humanoid robots exhibit higher degrees of freedom~\cite{cardenas2024xbg,fu2024humanplus,team2025gemini, bjorck2025gr00t} and more human-like motion, making trajectory quality crucial. 
To evaluate HumanPlus, Hu et al.~\cite{fu2024humanplus} divided the tasks into sub-tasks to assess the completion of short-term sequences of movements through success rate~\cite{fu2024humanplus}. In this way, the evaluation highlighted also what parts of the tasks were not performed. 
For human robot interaction tasks, however, the evaluation through success rate is inherently difficult, given the possible different ways of performing the same action. 
For instance, in payload handover scenarios, a robot may succeed in holding the object but still deviate significantly from the intended movement style (e.g., squeezing the object instead of supporting it naturally), potentially compromising generalization to different payloads.
%
%
%

Second, 
success rate evaluation suffers from limited reproducibility of results. To fairly compare trained model with success rate metrics, the methods need to be deployed on the robot and the experiments need to be executed under identical robot hardware, software, and human interaction conditions, which is notoriously difficult in HRIC contexts. 
The challenge of defining evaluation measures in HRIC applications has been highlighted in several works~\cite{coronado2022evaluating,kokotinis2023quantification,taghavi2021online,dietzel2024assessing,damacharla2018common}. Some studies reviewed general evaluation measures such as success rate, or efficiency~\cite{coronado2022evaluating,damacharla2018common}, while others defined task-specific scoring schemes~\cite{kokotinis2023quantification}. Other works~\cite{taghavi2021online,dietzel2024assessing} use trajectory similarity metrics based on Dynamic Time Warping (DTW)~\cite{muller2007dynamic} to compute the similarity between the human and the robot movements. However, DTW-based metrics are difficult to interpret, as small alignment cost variations can correspond to significant differences in movement quality.

To tackle the aforementioned limitations of evaluation schemes based upon success rate, we propose an automated way of evaluating the performance of models controlling the robot.  Our method evaluates behaviours based on the joint trajectories generated by the policies, supports reproducible testing of multiple policies on the same data, and allows assessment of model performance prior to robot deployment.

\begin{figure}[t!]
\begin{center}
\includegraphics[width=1.\linewidth]{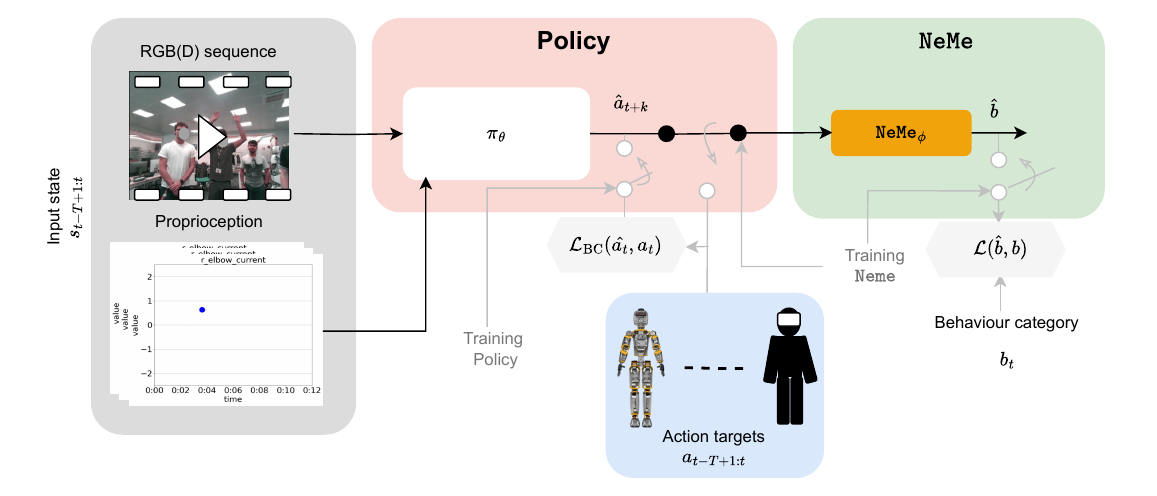}
\end{center}
\caption{\acronym{} is a neural network that classifies joint trajectories into behaviours and is employed as a meta-evaluator for learned imitation policies $\pi_{\theta}$.
}
\label{fig:framework}
\end{figure}

\section{The Neural Meta Evaluator (\acronym{})}~\label{sec:method} 
In this section, we present an automated approach for evaluating the performance of imitation learning (IL) policies. Our method evaluates behaviours based on the joint trajectories generated by the policies, supports reproducible testing of multiple policies on the same data, and allows assessment of model performance prior to robot deployment.

\textbf{Background \& Problem formulation.} Let us consider an Imitation Learning (IL) problem, where an autonomous robotic agent $\mathcal{R}$ processes $N$ demonstration trajectories  $\mathcal{D} = \{\tau_1, \dots, \tau_N\}$. The $i$-th trajectory is defined as a sequence $\tau_i = \{ (s_t, a_t, b_t)\}_{t=0}^{T_i}$,
where $s_t\in\mathcal{S}$ 
denotes the agent’s state at time $t$ (including perceptual observations such as RGB images or depth plus proprioceptive measurements like joint positions, which we denote with $j_t$), $a_t \in \mathcal{A}$ is the demonstrated action (e.g., target joint positions),\footnote{In our work, without loss of generality, we assume that target actions are provided by human teleoperation, and that a retargeting procedure is applied to adapt these actions to the agent’s embodiment.} $b_t$ is a semantic behaviour label (e.g., \textit{handshake}), and $T_i$ is the trajectory length. Each trajectory may have a different length. 
%
We model the agent’s policy as a parameterized function $\pi_\theta:\mathcal{S}^T\to\mathcal{A}$.  At time $t$, the policy produces $\hat a_t = \pi_\theta\bigl(s_{t-T+1:t}\bigr),$ i.e. the predicted action conditioned on the most recent $T$ states. 
The IL objective is to find,  over the demonstrations,  optimal parameters $\hat{\theta}$ of the policy $\pi$, which is usually implemented with a deep neural network. In literature, common approach to optimize the model are Behavioural Cloning  \cite{bain1995framework} and Diffusion-based objectives \cite{chi2023diffusionpolicy}. 
In this work, we focus on evaluating the \emph{quality} of the learned behaviours—namely the predicted trajectories $\{\hat a_t\}_{t=1}^{T_i}$—and their alignment with the semantic \textit{behaviour} labels $b_t$.
%
%
We define a \emph{behaviours} as the sequence of output joints positions in a time window of length $L$ i.e., the trajectory of the agent’s joints over time. A behaviour corresponds to a high-level action performed by the robot such as \emph{handwave} or \emph{handshake}, in reaction to the human performing the same movement. 

\textbf{The Neural Meta Evaluator.}
In order to assess the quality of trajectories generated by learned policies, we introduce the 
\textit{Neural Meta Evaluator} (\acronym{}--see Figure \ref{fig:framework}), 
a sequence modeling neural network designed to classify sequences of joints positions into behaviour categories. 
We train \acronym{} on teleoperation data, with the goal of learning to classify behaviours from 
human demonstrations. 
Subsequently, we leverage \acronym{} to objectively evaluate trajectories produced by IL policies. Indeed, \acronym{} is employed 
as a \textit{meta-evaluator}, i.e. a neural network 
that evaluates trajectories generated by learned policies. {Learning a behaviour classifier such as \acronym{} allows to learn behaviour trajectories directly from data, rather than assessing only the success/failure of an action, fostering a more comprehensive IL policy evaluation.}    

Formally, let the proprioceptive joint trajectory extracted from the $i$-th demonstration be defined as $\tau_i^J = \{ j_t \}_{t=0}^{L}$, where $L$ is the time-span of the demonstration and $j_t \in \mathcal{J}$ denotes the vector of joint states at time $t$.
Each demonstration is associated with a behaviour label $b_i \in \mathcal{B}$ (e.g., \emph{handwave}, \emph{handshake}). \acronym{}, parameterized by $\phi$, is trained to map the proprioceptive trajectory to its corresponding behaviour label, i.e. $E_{\phi}(\tau_i^J) \rightarrow \hat{b}_i,$ where $\hat{b}_i$ is the predicted behaviour label for the trajectory.

%
Training is performed by minimizing the cross-entropy loss over the demonstration dataset $\mathcal{D}^J = \{\tau_1^J, \dots, \tau_N^J\}$:

\[
\hat{\phi} = \arg\min_{\phi} \, \mathbb{E}_{\tau_i^J \sim \mathcal{D}} \left[ \mathcal{L}_{\mathrm{cls}} \Big( E_{\phi}(\tau_i^J), \, b_i \Big) \right],
\]

where $\mathcal{L}_{\mathrm{cls}}$ is the cross-entropy loss.
Once trained on teleoperation data, \acronym{} is employed as a \textit{meta-evaluator}—that is, a neural network trained independently that is subsequently used to assess the quality of trajectories generated by learned IL policies. Given a policy $\pi$ that produces a sequence of joint states $\hat{\tau}^a = \{ \hat{a}_t \}_{t=0}^{L},$ where $\hat a_t = \pi_\theta\bigl(s_{t-L+1:t}\bigr)$ represents the action predicted at time $t$,
\acronym{} predicts the behaviour associated with that trajectory as $\hat{b} = E_{\phi}(\hat{\tau}^a)$.

\textbf{Policy evaluation.} Since \acronym{} is trained to identify the behaviour in expert demonstrations at a trajectory level, 
we use it as performance measure to evaluate the 
joints trajectories predicted by the policies. 
The rationale behind \acronym{} is that the closer the trajectory produced by $\pi$ is to the characteristic patterns seen in the teleoperated data, the better its performance is evaluated.
Additionally, \acronym{} allows the offline evaluation of partial sequences 
comparing the predicted behaviour $\hat{b}$ with the annotated behaviour label $b$, hence computing 
a \textit{meta}-accuracy. Moreover, the evaluation through \acronym{} is performed before deploying the model on the robot and measuring the performance with success rate, thus avoiding the need to perform model selection on the physical robot. 
The offline evaluation, loading the collected visual input and proprioception, implies the assumption that the policy will predict behaviours similar to the collected ones. 
Since the models are trained to align to the target trajectories, this does not represent a limiting assumption, as our experimental campaign shows in the following. 

\section{Experiments}
\label{sec:exp}

\textbf{Robot platform.} 
Experiments were conducted with the ergoCub humanoid robot~\cite{ergocub2023}, a 150 cm tall, 56 DoF platform featuring fully articulated anthropomorphic hands.
Whole‐body motion is realized via a layered controller designed for bipedal locomotion~\cite{romualdiBenchmarkingDCMBased2020}, which accepts high‐level joystick‐like commands for the lower body and direct joint‐position targets for the upper body.
Following Cardenas et al. setup~\cite{cardenas2024xbg}, we consider the three joints in the neck, shoulder, elbow, torso, and wrists, the gaze tilt and the currents in the shoulder and elbow joints.

\begin{table}[t!] 
    \scriptsize
    \begin{center}
    \begin{tabular}{r l}
    \toprule
    \textbf{Task} & \textbf{Description} \\
    \midrule
    wave & Human greets the robot waiving both arms, the robot greets back raising one or both arms \\
    \rowcolor{gray!25} shake & Human extends the arm to offer the hand to the robot, the robot moves one hand to shake the human hand \\
    pick & \parbox{0.8\linewidth}{ Human approaches the robot from different angles and offers a payload to the robot, the robot extends both arms to receive the payload and then holds it until the human takes it back} \\
    \rowcolor{gray!25} walk & Human draws back from the robot moving both arms towards his body signalling the robot to walk and follow the person \\
    pick + walk & \parbox{0.8\linewidth}{ Human delivers a payload to the robot and signals to walk while carrying the payload (similarly to \textit{walk}), the robot walks towards the human balancing the payload without letting it fall} \\
    \rowcolor{gray!25} still & Human walks in front of the robot or does not trigger any movement, the robot is still and follows the human with the head \\
    pick + still & After \emph{ pick + walk}, the robot remains stationary holding the payload until further command \\
    \bottomrule
    \end{tabular}
    \end{center}
    \caption{Description of the tasks in the dataset.}
    \label{tab:task_description}
\end{table}

\begin{figure}
    \centering
    \includegraphics[width=0.95\linewidth,trim={5cm  13.5cm 0 0},clip]
    {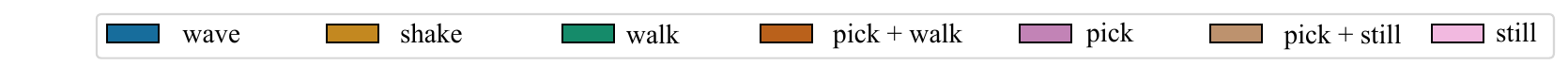} 
    \includegraphics[width=0.32\linewidth]{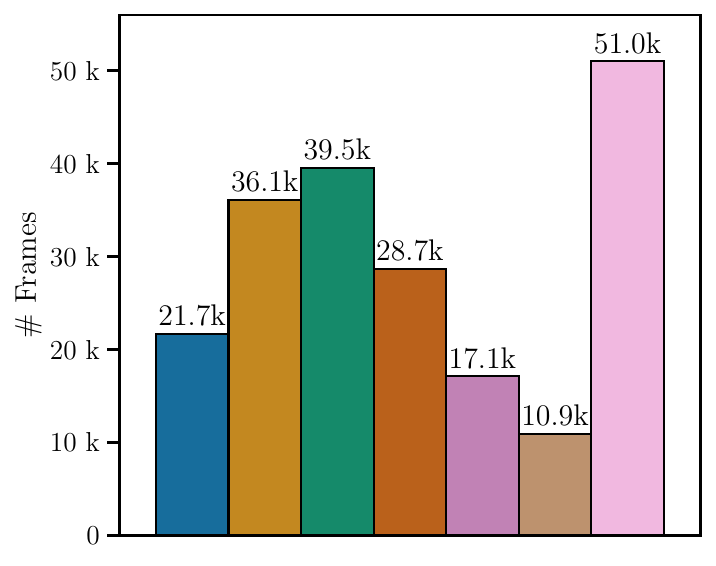}
    \includegraphics[width=0.32\linewidth]{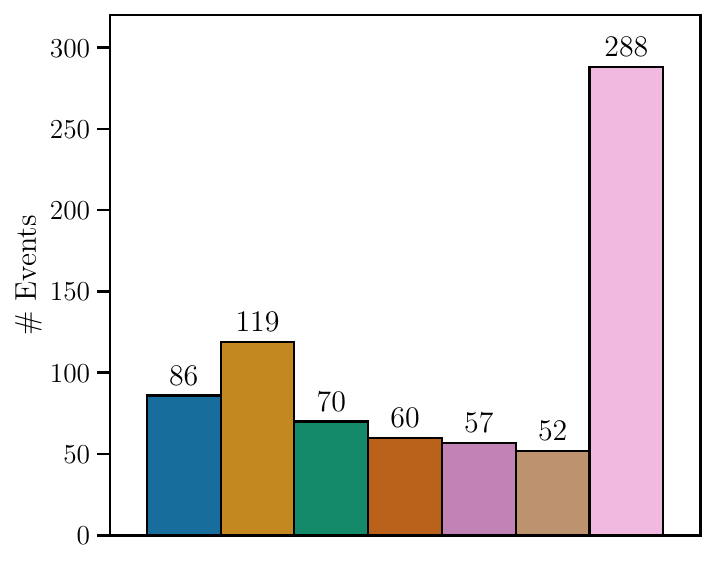}
    \includegraphics[width=0.32\linewidth]{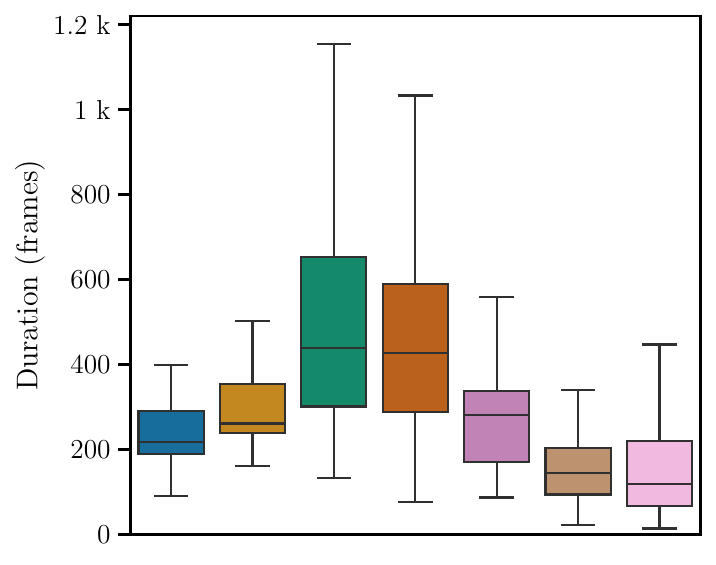} \\
    \caption{Statistical analysis of the  dataset. 
    We show the number of frames (1st column), number of events (2nd column), and the distribution of the frames duration (3rd column), per behaviour. 
    }
    \label{fig:data_stats}
\end{figure}

\textbf{Dataset.} 
We train and evaluate policies on an extended version (160 minutes) of the dataset introduced in~\cite{cardenas2024xbg} collected using human teleoperation through the avatar system~\cite{dafarra2024icub3} applied to the ergoCub platform: the human operator wears a suit sensorised with iFeel equipment~\cite{ifeel} that sends movement signals to the robot and a Virtual Reality helmet to perceive the environment through the robot vision system. 
The dataset consists of $640 \times 480$ RGB-D frames collected by the camera mounted in the robot head (Intel Realsense D450), robot joints state, and joints current~\cite{cardenas2024xbg}. 
During the demonstrations, the robot interacts with a human to perform 7 different tasks: 
\textit{handwave} (\textit{wave}), 
\textit{handshake}(\textit{shake}), 
\textit{walk}, 
\textit{walk with payload} (\textit{pick} + \textit{walk}), 
\textit{still with payload} (\textit{pick} + \textit{still}), 
\textit{payload reception} (\textit{pick}), 
 and \textit{standstill} (\textit{still})
(see definitions in Table~\ref{tab:task_description}).
The ground truth label of the behaviour has been manually associated to the corresponding frames in the dataset. 
Figure~\ref{fig:data_stats} summarizes, for each behaviour, both the total number of labeled frames (first plot) and the number of discrete events (second plot—i.e.\ the count of contiguous episodes of that behaviour) over the entire pool of recordings. To approximate a realistic deployment scenario, we retain the original training/validation split from Cárdenas \emph{et al.}~\cite{cardenas2024xbg}, but reserve all newly collected recordings for our test set. These additional test trials include human subjects not seen during training and were conducted under slightly different environmental conditions. This setup introduce a mild distribution shift in human subject appearance and interaction style, reflecting the variability and novelty that a real‐world user would present to the system. 

\textbf{Performance measures.} We evaluate the offline performance of the policies on the testing set using different indicators: meta-performances computed by comparing ground truth behaviour labels with the output of \acronym{} processing trajectories produced by policies, e.g.  meta-accuracy (mA) and \textit{per-class meta F1 score} (mF1), and \textit{per-class dynamic time warping}~\cite{muller2007dynamic,olsen2018simultaneous}. 
Additionally, we use \textit{Dynamic Time Warping} (DTW) to evaluate the similarity between trajectories predicted by trained policies and reference trajectories recorded during teleoperation as in previous HRIC works~\cite{taghavi2021online,dietzel2024assessing}. The DTW minimizes the cumulative Euclidean distance between the joint positions in the predicted trajectory and in the teleoperation one.
We compute the DTW for the actions \textit{handwave}, \textit{handshake}, \textit{payload reception}, and \textit{standstill} because the policies control only the joints of the robot upper body, while the joints in the legs are controlled using a walking signal, not the joint angle, hence their position is unknown. We divide the testing set into actions chunks (each chunk contains a single action) and we compute the average DTW over 21 3D positions on the body referred to the root link, and over all time instant in the action chunks.

\subsection{Training the Neural Meta Evaluator (\acronym{})}
In this section, we describe the first step of the proposed evaluation pipeline, i.e. training \acronym{} on target action trajectories. 

\textbf{Compared Models \& Training Details.} 
We compare the classification performance of several \acronym{} variants implemented using a variety of neural architectures for sequence modeling, drawn from the recent literature~\cite{tiezzi2024state}. We tested vanilla solutions for sequence classification--LSTMs~\cite{hochreiter1997long} and encoder-only Transformer~\cite{NIPS2017_3f5ee243} equipped with a \texttt{CLS} token, as well as modern evolutions (xLSTM~\cite{beckxlstm}) and  latest models from the deep state-space modeling literature, Mamba~\cite{gu2023mamba} and Mamba 2~\cite{dao2024transformers}, exploring their different contributions in term of computational efficiency, representational capability and long-context modelling.  For each architecture, we evaluated multiple combinations of values for the main hyper-parameters, by considering models with an amount of hidden layers $\in \{1,3,5\}$, each equipped with hidden layers with $h \in \{16,64,128\}$ neural units, learning rate $\beta \in \{10^{-2}, 10^{-3}, 10^{-4}\}$. We optimized the models for 30 epochs with the~\texttt{AdamW} \cite{loshchilov2017decoupled} optimization scheme, early stopping the training when the validation loss is not improving for 5 consecutive epochs. We perform data augmentation on the input trajectories to foster model generalization, with a Gaussian noise $\mathcal{N}(0, \sigma)$,  with $\sigma \in \{0, 10^{-1}, 10^{-2}\}$. In all the considered architectures, a projection layer maps the input features ($a_t \in \mathbb{R}^{24}$) to the hidden size $h$.  We selected the optimal values of the hyper-parameters from the aforementioned grids by maximizing their validation accuracy. We report the results averaged over three runs, performed with different initializations of the weights. We consider input temporal windows of different lengths $L \in \{ 16, 32, 64\}$ for the sequence model, which correspond to time windows of $1.6$, $3.2$, and $6.4$ seconds respectively. 

\begin{table}[t!] 
\small
    \begin{center}
    \begin{tabular}{l c c c c c c}
    \toprule
    \multirow{2}{*}{\textbf{Model}} &  \multicolumn{2}{c}{$L=16$} & \multicolumn{2}{c}{$L=32$} & \multicolumn{2}{c}{$L=64$} \\
    \cmidrule(l){2-3} \cmidrule(l){4-5} \cmidrule(l){6-7}
    & $A$ &  $F1$ & $A$ &  $F1$  & $A$ & $F1$ \\
    \midrule

\texttt{LSTM}~\cite{hochreiter1997long} &70.7 $\pm$ 1.7&65.9 $\pm$ 0.8&72.1 $\pm$ 1.3&66.6 $\pm$ 0.3&69.9 $\pm$ 3.9&65.0 $\pm$ 0.8\\
\texttt{Transf.}~\cite{NIPS2017_3f5ee243}&68.4 $\pm$ 1.5&61.5 $\pm$ 3.6&68.7 $\pm$ 1.2&61.6 $\pm$ 3.1&68.7 $\pm$ 2.4&61.8 $\pm$ 3.9\\
\texttt{xLSTM}~\cite{beckxlstm}&60.0 $\pm$ 5.2&54.1 $\pm$ 4.6&59.1 $\pm$ 3.6&53.3 $\pm$ 3.7&61.4 $\pm$ 6.0&55.4 $\pm$ 5.9\\
\texttt{Mamba}~\cite{gu2023mamba}&71.2 $\pm$ 3.8&62.8 $\pm$ 2.8&53.6 $\pm$ 13.8&51.9 $\pm$ 12.7&61.2 $\pm$ 0.1&56.6 $\pm$ 0.5\\
\texttt{Mamba2}~\cite{dao2024transformers}&63.3 $\pm$ 0.5&56.8 $\pm$ 2.7&58.1 $\pm$ 5.4&54.0 $\pm$ 4.9&56.2 $\pm$ 6.7&52.3 $\pm$ 5.4 \\
    \bottomrule
    \end{tabular}
\end{center}
    \caption{Behaviour classification with \acronym{}, test performance comparing different neural architectures.
    KEY -- $L$: input trajectory length, $A$: accuracy, $F1$: F1 score.}
    \label{tab:neural_eval_test_only}
\end{table}

\textbf{Results.} Table \ref{tab:neural_eval_test_only} compares the performance attained by the considered implementations. We report average accuracy ($A$) and $F1$ score over all the considered tasks (see Table \ref{tab:task_description}). Overall, the average scores lower than $73$ highlight that generalising the classification of behaviour is challenging. LSTM consistently performs well across all window sizes, with the highest accuracy and F1 score observed at $L = 32$. Notably, Mamba achieves higher performances at $L= 16$, but degrades the performance increasing $L$, suggesting sensitivity to sequence length. These results indicate that short time windows ($L = 32$) and classical models still provide a strong baseline for the considered data, generalising also to people unseen during training. 

\subsection{Evaluating the Neural Meta Evaluator (\acronym)}

We assess the effectiveness of the best trained Neural Meta Evaluator (LSTM with $L=32$) in two complementary model selection tasks: \textit{intra-policy evaluation}, where \acronym{} is employed to select the best performing model, %
trained using BC, over different training epochs,  
and \textit{inter-policies evaluation}, in which \acronym{} compares the performance of different architectures trained with BC. 

\textbf{Compared policies.}
We evaluate the performance of different versions of Exteroceptive Behaviour Generation~\cite{cardenas2024xbg}, a policy designed to work in HRIC applications: using RGB and depth input (\texttt{XBG}), using only RGB information (\texttt{XBG-RGB}). 
\texttt{XBG}~\cite{cardenas2024xbg}  feeds an LSTM~\cite{hochreiter1997long} with the concatenation of RGB, depth and proprioception embeddings of an input window. The model predicts joints positions corresponding to a single time step in the future.  
Starting from XBG~\cite{cardenas2024xbg}, we implement \texttt{XBG-RGB} by removing the depth input and the corresponding encoder. \texttt{XBG} and \texttt{XBG-RGB} are trained using the same setup~\cite{cardenas2024xbg}: 12 epochs, batch size 64, 2 NVIDIA A100 gpus, learning rate $\beta=5*10^{-4}$, Adam as optimizer with default parameters, an input sequence of 16 frames at 10 Hz. The policy learns to predict the joint positions at the subsequent tenth frame. We crop the frames symmetrically on the longest side and resize it to $224 \times 224$, and we apply with a probability of 0.3 the same augmentations in Cardenas et al.'s setup~\cite{cardenas2024xbg}.

\textbf{Success rate protocol.} We consider 3 people interacting with the robot, performing 10 trials each. In particular, we consider each action a success if the robot reacts within 3 seconds, and if after performing an action it returns back to the original pose, within 3 seconds. Additionally, for the \textit{receive payload} task, the robot should keep the payload up for 5 seconds. If these conditions are not met or if the robot performs a task different from the one requested by the human, then it is considered a failure. Note that SR test are open-world tests, while policy evaluation is performed with \acronym{} on the test set described above, which cannot be used here since does not allow the robot to actually interact.

\textbf{Intra-policy evaluation results.}
Figure~\ref{fig:main_figure} shows the performance of different methods to select the best performing weights of the policy, over the 12 epochs. In particular, we considered to select the model based on 4 criteria: loss on the validation split of the dataset (the lower, the better), the DTW value (the lower, the better) on the testing split of the dataset,  the success rate (the higher, the better) of each training epoch deployed and tested on the robot, and the mF1 score (the higher, the better) obtained by the \acronym{} evaluation pipeline, computed on the testing split of the dataset. Selecting the best weights using the validation loss leads to chose the model trained for one epoch, that however corresponds to a low average success rate (20\%). The model achieves the lowest DTW at the last epoch, where the success rate has decreased from the peak and is around 65\%. On the contrary, leveraging  the mF1 score of \acronym{} to select the best model weights, the ones corresponding to the 8th epoch would be selected--that is closer to the peak of success rate compared to other measures. The standard deviation of the success rate shows that the metric has high variability across human demonstrators, due to the variability in testing conditions (different person interacting with the robot, different backgrund or illumination condition) and the subjectiveness biasing human evaluators. In this sense, \acronym{} offers an offline alternative for model selection, yielding predictions that better align with the best-performing models according to SR, without requiring interaction with the robot or incurring variability-related uncertainty. 
 


\textbf{Inter-policies evaluation results.} In this case, we apply the best-performing evaluator to select the optimal training epoch for the two considered imitation learning models tailored to the platform, \texttt{XBG} and \texttt{XBG-RGB} (weight corresponding to the 8th and 4th epochs, respectively). 
As reported in Table~\ref{tab:imitation_learning_experiment_full}, \acronym{} assigns an higher average mF1 score to trajectories generated by \texttt{XBG} (71.3) compared to the ones generated by \texttt{XBG-RGB} (69.9). This result is consistent with the actual SR measured on the robot (70.0 vs 61.7), suggesting that \acronym{}’s evaluations reflect real-world performance.  
These results support \acronym{}'s effectiveness in cross-architecture model selection, enabling reliable comparisons without the cost and variability of on-robot deployment.

\begin{table}[t!] 
    \scriptsize
    \setlength{\tabcolsep}{2.2pt} 
    \begin{center}
    \begin{tabular}{l cccc cccc cccc cccc cccc}
    \toprule
    \textbf{Method} & \multicolumn{4}{c}{\textbf{Handwave}} &  \multicolumn{4}{c}{\textbf{Handshake}} &  \multicolumn{4}{c}{\textbf{Payload reception}} & \multicolumn{4}{c}{\textbf{Standstill}} & \multicolumn{4}{c}{\textbf{Average}} \\
    \cmidrule(l){2-5} \cmidrule(l){6-9} \cmidrule(l){10-13} \cmidrule(l){14-17} \cmidrule(l){18-21}
    & mA & mF1 & $D$ & $S$ & mA & $F1$ & $D$ & $S$ & mA & mF1 & $D$ & $S$ & mA & mF1 & $D$ & $S$ & mA & mF1 & $D$ & $S$ \\
    \midrule
    XBG~\cite{cardenas2024xbg}  & 84.1 & 77.9 & 2.29 & 73.3 & 89.6 & 81.9 & 3.13 & 46.7 & 90.6 & 81.7 & 1.86 & 83.3 & 29.6 & 43.5 & 1.44 & 76.7 & 73.5 & 71.3 & 2.18 & 70.0 \\
    XBG-RGB~\cite{cardenas2024xbg} & 86.3 & 82.6 & 2.73 & 73.3 & 80.1 & 79.4 & 3.34 & 13.3 & 81.2 & 77.6 & 2.09 & 86.7 & 26.4 & 40.0 & 1.43 & 73.3 & 68.5 & 69.9 & 2.40 & 61.7 \\
    \bottomrule
    \end{tabular}
    \end{center}
    \caption{Performance comparison of the imitation learning models measured through the Neural Evaluator implemented as an LSTM with $L=32$. KEYS -- mA: meta-accuracy, mF1: meta-F1 score, $D$: dynamic time warping distance, $S$: success rate.} \label{tab:imitation_learning_experiment_full}
\end{table}

\section{Conclusion}
\label{sec:conclusion}
In this paper we proposed \acronym{}, a Neural Meta Evaluator for offline  performance assessment of imitation learning policies
. \acronym{} is trained to classify robot behaviours directly from joints trajectories. Then, it is employed as a meta-evaluator to evaluate the trajectories produces by learned policies. We showed that generalising to unseen trajectories is challenging, due to the variations in robot movements. With \acronym{} we are capable to select the best performing weights across different epochs, further showing that compared to using Dynamic Time Warping and validation loss, our method is the most aligned with the selection based on the average success rate; \acronym{} allows also to compare different policies and to select the best performing one, aligning with those achieving the highest success rate on the robot, indicating its reliability as a surrogate for deployment performances. 
As future works, we will focus on applying the neural evaluator approach to other policies, increasing the number of human-robot interaction actions, and considering also the application to robotic arms.



\section{Limitations}

The main limitation of our work is the assumption that, during testing, the robot behaves in a similar way compared to the collected samples. This limitation is due to the fact that we feed the vision and the proprioception signals of the testing samples into the robot policy, even if the robot joints moves in a different way. However, the different joint trajectories should cause the neural evaluator to mis-classify the behaviour, hence impacting the overall performance of the policy.  
The second limitation concerns the trained neural evaluator. In case the robot platform or the actions in the dataset change, \acronym{} needs retraining to learn a new set of actions or the values of the joints positions relative to the new platform. However, re-training the neural evaluator avoids the time consuming activity of deploying and testing each trained policy on the robot. 



\bibliography{biblio}  

\end{document}